\documentclass[conference]{IEEEtran}
\IEEEoverridecommandlockouts
\usepackage{cite}
\usepackage{amsmath,amssymb,amsfonts}
\usepackage{algorithmic}
\usepackage{graphicx}
\usepackage{textcomp}
\usepackage{xcolor}
\usepackage[binary-units]{siunitx}
\usepackage{url}  
\def\BibTeX{{\rm B\kern-.05em{\sc i\kern-.025em b}\kern-.08em
    T\kern-.1667em\lower.7ex\hbox{E}\kern-.125emX}}
\begin{document}
\bstctlcite{IEEEexample:BSTcontrol}
\setlength{\textfloatsep}{1cm}

\title{Demo: LE3D: A Privacy-preserving Lightweight Data Drift Detection Framework}

\author{\IEEEauthorblockN{Ioannis Mavromatis and Aftab Khan}
\IEEEauthorblockA{Bristol Research and Innovation Laboratory (BRIL), Toshiba Europe Ltd., Bristol, UK \\
Emails: \{Ioannis.Mavromatis, Aftab.Khan\}@toshiba-bril.com}}

\maketitle

\begin{abstract}
This paper presents LE3D; a novel data drift detection framework for preserving data integrity and confidentiality. LE3D is a generalisable platform for evaluating novel drift detection mechanisms within the Internet of Things (IoT) sensor deployments. Our framework operates in a distributed manner, preserving data privacy while still being adaptable to new sensors with minimal online reconfiguration. Our framework currently supports multiple drift estimators for time-series IoT data and can easily be extended to accommodate new data types and drift detection mechanisms. This demo will illustrate the functionality of LE3D under a real-world-like scenario.
\end{abstract}


\vspace{-1mm}

\section{Introduction and Motivation}
\vspace{-1mm}

IoT sensors are found in numerous domains, e.g., air pollution monitoring, farming, smart cities, etc.~\cite{IoTbasedSmartCities}. These applications rely on data fidelity. Considering the scale and economic viability, the use of low-cost sensors is inevitable. However, the low-cost nature of these sensors, the differences between manufacturers, the lack of reliable calibration, and the ``silicon gamble'', can lead to inconsistencies. The differences become even more prominent when data from different devices are compared~\cite{sensorComparison}, making the data's relative measurements the only viable strategy for comparison between them.

Moreover, the data integrity in the above applications can be of paramount importance~\cite{dataIntegrity}. Altered data streams, either from benign cases (e.g., faulty sensors) or malicious actions (e.g., unauthorised data tampering), can disrupt or bias an application and result in widespread damage and outages. Therefore, preserving data privacy while ensuring their integrity has recently become a big topic of scientific discussion, with various data drift solutions being proposed, e.g.,~\cite{mlDataDrift,mlAutoencoder,PWPAE}. 

Previous activities either focused on improving the prediction accuracy, operated without considering data privacy (where the data is stored and processed), or without provision for real-world deployment. Inspired by the above, we present \textit{Lightweight Ensemble of Data Drift Detectors (LE3D)}, 
a novel lightweight data drift detection framework. Its operation is two-fold, i.e., it can be used: \textit{1)} for evaluating novel data drift detection strategies, and \textit{2)} for real-world deployments detecting different types of drift in multiple sensors. 
Our framework is publicly available at {\tt\small github.com/toshiba-bril/le3dDataDriftDetector}.

\section{LE3D: Main System Components}\label{sec:maincomponents}
\vspace{-1mm}

LE3D is designed with both research and real-world scenarios - consumer applications in mind. It consists of an extensible drift detection framework and a set of supporting tools. Within LE3D, an \textit{estimator} can be a statistical or a Machine Learning process that classifies a sensor sample as drifting or not. The \textit{estimators} can operate in an \textit{online learning}-fashion, adapting to the data distribution changes or working as static classifiers (e.g., thresholding methods). LE3D supports various statistical drift estimators, i.e., ADaptive WINdowing (ADWIN), Page-Hinkley Test (PHT), and Kolmogorov-Smirnov Windowing (KSWIN) for time series data~\cite{estimators}. These estimators or the data fed into them can be easily replaced or extended as required.

A \textit{detector} plays multiple roles. Firstly, it handles received data streams. Later, based on the data type, it can assign one or many \textit{estimators} for each sensor stream. Ensemble strategies can also be introduced to enhance the classification accuracy (e.g., voting mechanisms taking into account windows of samples and multiple \textit{estimators}' outcome). Finally, if required by an end-user, a \textit{detector} can relay the sensor data to the backend for visualisation, storage, and demonstration purposes. 

LE3D ensures data privacy. All decisions are taken at the ``edge'' without data being exchanged across the network. Multiple \textit{detectors} can collectively classify the detected drift type as \textit{natural} -- happening concurrently across multiple sensors -- or \textit{abnormal} -- only one sensor is drifting. This is performed through an \textit{aggregator} that operates in parallel with each \textit{detector}. The \textit{aggregators} share only the classification decision, the metadata provided by an endpoint, and the result of a one-sample K-S test, with the surrounding devices preserving the data privacy. More details about the algorithms and intelligence introduced within LE3D can be found in~\cite{le3d_2022}.

LE3D comes with a set of supporting tools for visualisation and experimentation. As access to real-world drifting endpoints is not always possible, LE3D provides a \textit{streamer} and an \textit{emulator}. The \textit{streamer} streams ``real-world'' data from a pre-existing dataset (in CSV format). On the other hand, an \textit{emulator} generates ``realistic'' emulated data streams and drifts on demand. Their statistical properties can be based on real data to ensure realistically emulated sensors. Moreover, a \textit{matching} framework ensures the system's scalability when large-scale experiments are conducted. This framework operates as an oracle and associates different endpoints, \textit{emulators}, \textit{streamers}, and \textit{detectors} for large-scale experimentation. Finally, LE3D comes with a simple GUI for visualising and controlling the drifts introduced. The GUI can be easily extended to support more drifts or visualisation interfaces.

\begin{figure*}[t]
    \centering
    \includegraphics[width=1\textwidth]{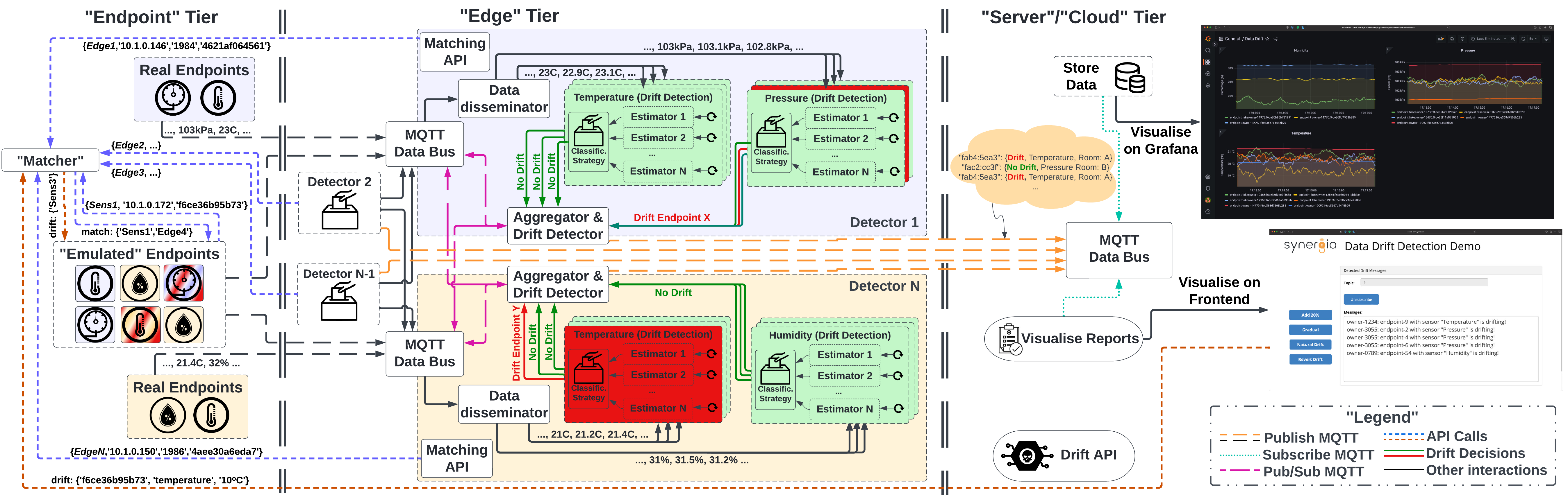}
    \vspace{-7mm}
    \caption{A detailed system diagram visualising the different system components and the interactions between them. LE3D operates in a three-tier architecture, with each component running as a microservice. All interactions occur either via well-defined MQTT topics and messages or RESTful APIs.}

    \label{fig:highleveldiagram}
\end{figure*}

\section{LE3D: System Architecture/Implementation}

LE3D operates in a standard three-tier architecture (Fig.~\ref{fig:highleveldiagram}): cloud, edge, and endpoint. The ``cloud'' hosts the frontend, the \textit{matching} framework, and a database for storage. The ``edge'' tier hosts the \textit{detectors}, the data messaging bus, the \textit{aggregators}, and is responsible for sharing the results with the ``cloud''. Finally, the ``endpoints'' can be either real or emulated sensor streams incorporating no intelligence.

All the different components are independent microservices that can be deployed and updated on demand. LE3D is highly scalable and extensible and allows the end-user to implement new functions within the existing applications or replace them entirely with new ones (e.g., replace the frontend GUI with a different application). LE3D can be deployed locally for testing (on a single computer) or across a distributed Kubernetes cluster. At its minimum configuration, LE3D requires just a single \textit{detector} on each ``edge'' device and a way to receive sensor data from one or multiple endpoints. Our system architecture diagram can be found in Fig.~\ref{fig:highleveldiagram}

The communication plane is based on MQTT and RESTful APIs. All the sensor data and decisions are exchanged via multiple MQTT predefined topics. More specifically, the detector decisions are published as retained messages, while the sensor samples as regular ones. Moreover, the interactions between the applications (e.g., the association of emulators to detectors) are done via RESTful APIs. 

The default configuration can be overridden with environmental variables during the execution. All the above results in a very flexible, robust, and extensible implementation that can be used either for research-driven drift detection activities or deployment in realistic scenarios. Moreover, the core functionality of LE3D introduces minimal overhead and can be scaled up across tens or hundreds of devices. 

Our framework was implemented in Python 3.9.12. The existing estimators' functionality is based on River online/streaming ML package~\cite{river}. All statistical calculations and optimisation mechanisms are based on SciPy and NumPy libraries. The MQTT messaging relies on Eclipse's Paho MQTT client implementation. 
The frontend and all the RESTful APIs are developed with Flask. The functionality of the different components described in Sec.~\ref{sec:maincomponents} was developed in-house. Finally, the \textit{detector}'s and \textit{aggregator}'s functionality has been tested on a Raspberry Pi (RPi) Compute Module 3b+~\cite{le3d_2022}, with a BCM2837B0 Cortex-A53 64-bit \SI{1.2}{\giga\hertz} System-on-a-Chip (SoC) and \SI{1}{\giga\byte} of RAM. This RPi was chosen as a representative resource-constrained IoT device.


\section{Demonstration}

The demonstration will showcase the functionality of LE3D in a real-world-like scenario. Various emulated endpoints and streamers will be executed, generating realistic traffic targeting multiple detectors. An end-user will introduce various drifts, which will be identified by a lightweight ensemble data drift detection implementation running in a distributed fashion. Finally, all the results will be displayed to the end user.


\vspace{-1mm}

\bibliographystyle{IEEEtran}
\bibliography{IEEEabrv,bib.bib}

\begin{thebibliography}{1}
\providecommand{\url}[1]{#1}
\csname url@samestyle\endcsname
\providecommand{\newblock}{\relax}
\providecommand{\bibinfo}[2]{#2}
\providecommand{\BIBentrySTDinterwordspacing}{\spaceskip=0pt\relax}
\providecommand{\BIBentryALTinterwordstretchfactor}{4}
\providecommand{\BIBentryALTinterwordspacing}{\spaceskip=\fontdimen2\font plus
\BIBentryALTinterwordstretchfactor\fontdimen3\font minus
  \fontdimen4\font\relax}
\providecommand{\BIBforeignlanguage}[2]{{%
\expandafter\ifx\csname l@#1\endcsname\relax
\typeout{** WARNING: IEEEtran.bst: No hyphenation pattern has been}%
\typeout{** loaded for the language `#1'. Using the pattern for}%
\typeout{** the default language instead.}%
\else
\language=\csname l@#1\endcsname
\fi
#2}}
\providecommand{\BIBdecl}{\relax}
\BIBdecl

\bibitem{IoTbasedSmartCities}
H.~Arasteh, V.~Hosseinnezhad \emph{et~al.}, ``{IoT-based Smart Cities: A
  Survey},'' in \emph{Proc. of IEEE EEEIC 2016}, Jun. 2016, pp. 1--6.

\bibitem{sensorComparison}
P.~Ferrer-Cid, J.~M. Barcelo-Ordinas \emph{et~al.}, ``{A Comparative Study of
  Calibration Methods for Low-Cost Ozone Sensors in IoT Platforms},''
  \emph{{IEEE} Internet Things J.}, vol.~6, no.~6, pp. 9563--9571, Jul. 2019.

\bibitem{dataIntegrity}
M.~N. Aman, B.~Sikdar \emph{et~al.}, ``{Low Power Data Integrity in IoT
  Systems},'' \emph{{IEEE} Internet Things J.}, vol.~5, no.~4, May 2018.

\bibitem{mlDataDrift}
O.~A. Wahab, ``{Intrusion Detection in the IoT under Data and Concept Drifts:
  Online Deep Learning Approach},'' \emph{{IEEE} Internet Things J.}, pp. 1--1,
  Apr. 2022.

\bibitem{mlAutoencoder}
B.~Friedrich, T.~Sawabe \emph{et~al.}, ``{Unsupervised Statistical Concept
  Drift Detection for Behaviour Abnormality Detection},'' \emph{Applied
  Intelligence}, vol.~58, no.~3, pp. 509--523, May 2022.

\bibitem{PWPAE}
L.~Yang, D.~M. Manias \emph{et~al.}, ``{PWPAE: An Ensemble Framework for
  Concept Drift Adaptation in IoT Data Streams},'' in \emph{Proc. of IEEE
  GLOBECOM 2021}, Dec. 2021, pp. 01--06.

\bibitem{estimators}
F.~Bayram, B.~S. Ahmed \emph{et~al.}, ``{From Concept Drift to Model
  Degradation: An Overview on Performance-aware Drift Detectors},''
  \emph{Knowledge-Based Systems}, vol. 245, p. 108632, Mar. 2022.

\bibitem{le3d_2022}
I.~Mavromatis, A.~S{\'{a}}nchez{-}Momp{\'{o}} \emph{et~al.}, ``{LE3D: A
  Lightweight Ensemble Framework of Data Drift Detectors for
  Resource-Constrained Devices},'' \emph{arXiv:2211.01840 [cs.LG]}, Jan. 2023.

\bibitem{river}
J.~Montiel, M.~Halford \emph{et~al.}, ``{River: Machine Learning for Streaming
  Data in Python},'' \emph{J. Mach. Learn. Res.}, vol.~22, no.~1, Jul. 2022.

\end{thebibliography}

\end{document}